# ViTs are Everywhere: A Comprehensive Study Showcasing Vision Transformers in Different Domain

Md Sohag Mia*
*School of Artificial Intelligence Nanjing University of Information Science and Technology*
Nanjing, China
shuvosohagahmmed@gmail.com

Abu Bakor Hayat Arnob*
*School of Artificial Intelligence Nanjing University of Information Science and Technology*
Nanjing, China
satcarnob985@gmail.com

Abdu Naim+
*School of Artificial Intelligence Nanjing University of Information Science and Technology*
Nanjing, China
naimabdu@nuist.edu.cn

Abdullah Al Bary Voban+
*School of Artificial Intelligence Nanjing University of Information Science and Technology*
Nanjing, China
voban@nuist.edu.cn

Md Shariful Islam
*Beijing Institute of Technology*
Beijing, China
sharif@bit.edu.cn

*Abstract*—**Transformer design is the de facto standard for natural language processing tasks. The success of the transformer design in natural language processing has lately piqued the interest of researchers in the domain of computer vision. When compared to Convolutional Neural Networks (CNNs), Vision Transformers (ViTs) are becoming more popular and dominant solutions for many vision problems. Transformer-based models outperform other types of networks, such as convolutional and recurrent neural networks, in a range of visual benchmarks. We evaluate various vision transformer models in this work by dividing them into distinct jobs and examining their benefits and drawbacks. ViTs can overcome several possible difficulties with convolutional neural networks (CNNs). The goal of this survey is to show the first use of ViTs in CV. In the first phase, we categorize various CV applications where ViTs are appropriate. Image classification, object identification, image segmentation, video transformer, image denoising, and NAS are all CV applications. Our next step will be to analyze the state-of-the-art in each area and identify the models that are currently available. In addition, we outline numerous open research difficulties as well as prospective research possibilities.**

## I. INTRODUCTION

Deep neural networks (DNNs) have evolved into the foundation of artificial intelligence (AI) systems at present. Different sorts of networks have traditionally been used for different types of activities. The classical form of the neural network, for example, is the multi-layer perceptron (MLP) or fully connected (FC) network, which is made up of numerous linear layers and nonlinear activations layered together [1]. Convolutional neural networks (CNNs) are neural networks that use convolutional layers and pooling layers to handle shift-invariant input such as images [2]. Furthermore, recurrent neural networks (RNNs) use recurrent cells to handle sequential or time series input [3]. Transformers are a new sort of neural network. It primarily employs the self-attention mechanism [4] to extract intrinsic properties and has significant potential for widespread adoption in AI applications. Transformer was initially used for natural language processing (NLP) tasks, where it significantly improved performance [5]. For example, Vaswani et al. [6] first proposed a transformer based on an attention approach, for machine translation and English constituency parsing tasks. Devlin et al. [5] presented a novel language representation model called BERT (short for Bidirectional Encoder Representations from Transformers), which pre-trains a transformer on the unlabeled text while taking into consideration the context of each word because it is bidirectional. BERT achieved state-of-the-art performance on 11 NLP tasks when it was published. Brown et al. [7] used 175 billion parameters to train a huge transformer-based model called GPT-3 (short for Generative Pre-trained Transformer 3) using 45TB of compressed plaintext data it delivered strong results on a variety of downstream natural language tasks without the need for fine-tuning. With their high representation capacity, transformer-based models have made important advances in NLP. To enable future research on various themes, We group transformer models according to their application situations, as shown in Table 1. Backbone networks, high/mid-level vision, low-level vision, and multimodality are the primary categories. From an architectural standpoint, the backbone of ViTs is solely composed of self-attention mechanisms, which have demonstrated extraordinary performance in visual tasks. Two primary notions have greatly influenced the evolution of ViT models. (a)A self-attention method in which ViTs collected long-range token relationships in a global context, in the same way, that typical recurrent neural networks do. (b) The second significant advantage is the ability to train on large-scale unlabeled datasets and fine-tune them on other tasks using tiny datasets. Fig. 2 [8] depicts the vision transformer's development chronology. There will be many more landmarks in the future.

Table 1:Recent works of Vision Transformers (ViTs)

| Category | Sub-category | Method | *Highlights* | Publication |
|---|---|---|---|---|
| Backbone | Supervised | ViT [62] | Image Patches, Standard Transformer | ICLR 2021 |
| | | Swin [19] | Shifted Window, Window Based Self-attention | ICCV 2021 |
| High/Mid-Level Vision | Object Detection | DETR [14] | Set-based prediction, bipartite matching | ECCV 2020 |
| | | Deformable DETR [91] | Deformable | ICLR 2021 |
| | Segmentation | Max-DeepLab [92] | PQ-style bipartite matching, dual-path transformer | CVPR 2021 |
| | Pose-Estimation | METRO [93] | Progressive Dimensionality reduction | CVPR 2021 |
| Low -Level Vision | Image-generation | TransGAN | Transformer based GAN | NeurIPS 2021 |
| Multimodality | classification | CLIP | NLP supervision for images | arXiv 2021 |
| | Image-generation | DALL-E | Zero-shot text -to image generation | ICML 2021 |

## II. RELATED WORKS

Several surveys on ViTs have been undertaken in the literature. [9] examines the transformer's theoretical ideas, base, and applications for memory efficiency. They also talked about how efficient transformers may be used in NLP. CV duties, on the other hand, were not included. A similar work [10] investigated the theoretical features of the ViTs, transformer foundations, the function of multi-head attention in transformers, and transformer applications in image classification, segmentation, super-resolution, and object identification. The investigation excluded transformer applications for picture compression. The authors of [11] described transformer topologies for segmenting, classifying, and detecting objects in images. This survey excluded activities related to CV and image processing such as image super-resolution, denoising, and compression. The authors of [12] describe several transformer topologies for computational visual media. The authors highlighted how transformers may be used for low-level vision and generation tasks such as image colorization, super-resolution, image production, and text-to-image conversion. Furthermore, the study concentrated on high-level vision tasks including segmentation and object recognition. super-resolution, image production, and text-to-image super-resolution, image production, and text-to-image conversion.

While other reviews [8]–[13]focused on a narrower filed, which predominantly honed in on domain-specific investigations, our survey adopts a more expansive approach. Our intention is to provide a panoramic view that encompasses a wide spectrum of domains, thus demonstrating the versatile adaptability of Vision Transformers (ViTs) and their remarkable performance across an array of applications. By undertaking this comprehensive exploration, we aim to paint a more detailed and illuminating picture of the current landscape in machine learning. This comprehensive perspective not only serves to illuminate the multifaceted capabilities of ViTs but also empowers researchers with invaluable insights for selecting the most suitable approaches tailored to their specific research needs. In essence, our survey seeks to offer a holistic understanding of the ViT paradigm, shedding light on its myriad possibilities and paving the way for more informed and strategic research directions in the field of machine learning.

## III. OVERVIEW OF VISION TRANSFORMER

In The compilation of various research, diverse advancements in object detection and self-supervised learning with transformer architectures are presented. The DETR[14] approach redefines object detection as a set prediction problem, employing transformer encoders and decoders to simplify the detection pipeline and enhance predictions. Similarly, Swin Transformers are leveraged for self-supervised learning, combining elements from MoCo v2[15] and BYOL[16] to achieve remarkable accuracy on ImageNet-1K evaluations. Cross-modality transfer learning is explored, highlighting the effectiveness of Convolutional Neural Networks and Vision Transformers in bridging the gap between 2D and 3D vision. The innovation continues with simCrossTrans[17], a simple yet highly effective cross-modality transfer learning technique for object detection. Additionally, the adaptation of Vision Transformer backbones for object detection eliminates the need for complex redesigns, presenting a promising direction for future research. MIMDET[18] further refines the use of vanilla ViTs for object detection, outperforming hierarchical counterparts on benchmark datasets. The realm of open-vocabulary object detection is expanded through region-aware pretraining, resulting in competitive zero-shot transfer detection and state-of-the-art performance on LVIS benchmarks. These collective contributions drive the evolution of object detection techniques and underline the potential of transformer-based models in various domains.

In a convergence of transformative advancements in computer vision and language processing, several innovative approaches stand out. The Swin Transformer[19] introduces a hierarchical architecture with shifted windows to address discrepancies between language and vision domains, effectively modeling multi-scale features and achieving linear computational efficiency with image size. In CrossViT,[20] a dual-branch converter is proposed to enhance image feature representation by aggregating patches of varying sizes, yielding competitive performance against traditional convolutional neural networks. Robustness gains new ground in Towards Robust Vision Transformer (RVT),[21] showcasing improved resilience and generalization across ImageNet and robustness benchmarks, with RVT-S leading robustness leaderboards. The CSWin Transformer[22] pioneers the CrossShaped Window self-attention mechanism, optimizing the trade-off between spatial interaction modeling and computation costs, achieving high accuracy and segmentation performance. Additionally, Global Context Vision Transformers ingeniously integrate global context self-attention modules with local self-attention, efficiently capturing spatial interactions of varying ranges. In the realm of hybrid architectures, FastViT[23] presents a groundbreaking solution, combining transformer and convolutional designs to yield an unparalleled latency-accuracy trade-off. This innovation surpasses state-of-the-art models in speed while outperforming MobileOne's[24] ImageNet accuracy by 4.2%. These paradigm-shifting approaches collectively redefine the landscape of computer vision by synergizing the capabilities of transformers and convolutional architectures.

In a dynamic landscape of medical image analysis, transformative breakthroughs are reshaping the field. Leveraging the Vision Transformer model with ConvNets, ViT-V-Net[25] achieves substantial progress in volumetric medical image registration, adeptly handling long-range spatial relations. The authors of [26] tackles weakly supervised classification for whole slide image-based pathology diagnosis, offering a potent solution through TransMIL's[26] effective handling of balanced/unbalanced and binary/multiple classifications with strong interpretability. UNetFormer[27] introduces an adaptable paradigm with a 3D Swin Transformer encoder and CNN-based decoders, offering versatile accuracy-computation trade-offs. SEViT[28] fortifies Vision Transformers (ViTs) against adversarial attacks, notably in gray-box environments, through a Self-Ensembling methodology validated in chest X-ray and fundoscopy experiments. Pioneering consistency-aware pseudo-label-based self-ensemble, combining Vision Transformers and CNNs, advances performance significantly. MEW-UNet[29] stands out in medical image segmentation, outperforming by 10.15 mm on the Synapse dataset. A hybrid MedViT[30] model capitalizes on the global interconnectedness of Vision Transformers and the localization capabilities of local CNNs, exemplifying robustness and generalization. MDViT's[30] introduction underscores enhanced medical image segmentation through domain adapters, bolstering representation learning across diverse domains. Lastly, SEDA's innovative approach utilizes defensive distillation and adversarial training to bolster tuberculosis categorization from chest X-rays, harnessing CNN blocks and spatial features for heightened efficiency. These pioneering methodologies collectively redefine medical image analysis and prognosis, fostering transformative advancements at the nexus of artificial intelligence and healthcare. In the ever-evolving landscape of image restoration and generative modeling, pioneering methodologies have emerged. NOISE2VOID (N2V)[31] presents a unique training strategy for denoising biological imaging data, demonstrating moderate denoising performance without relying on noisy image pairs or clean targets. SwinIR[32] introduces the Swin Transformer-based SwinIR[32] baseline model, revolutionizing image restoration and surpassing state-of-the-art methods while significantly reducing parameter count. The groundbreaking Few-Shot Diffusion Models (FSDM) framework empowers few-shot generation through conditional DDPMs, facilitating sample generation for new classes and datasets while enhancing training convergence. The innovative continuous Wavelet Sliding-Transformer (DnSwin)[33] addresses frequency dependencies in image recovery, excelling in noise removal and information recovery. GenViT[34] fuses Vision Transformer and Diffusion Denoising Probability Models to create a hybrid Generative ViT (GenViT)[34], outperforming predecessors in both generative and discriminative tasks, thereby extending its application horizon. A novel Dual branch Deformable Transformer (DDT)[35] denoising network advances local-global interactions, prioritizing crucial regions and yielding superior performance with optimized computational costs. These transformative approaches collectively shape the frontier of image restoration, generative modeling, and denoising

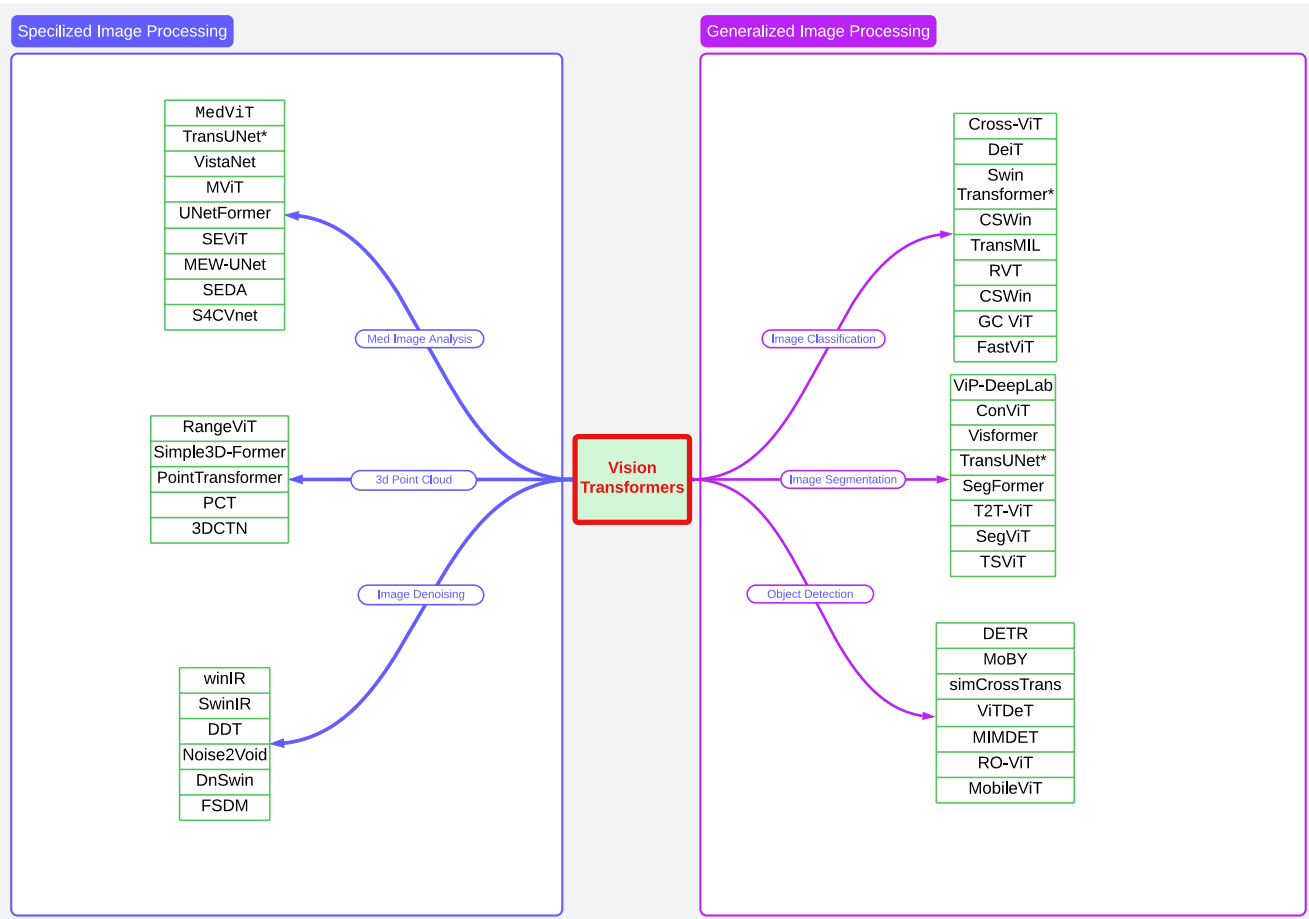

Fig. 1. Vision Transformers in various fields.

techniques, propelling the field towards unprecedented horizons.

In the realm of 3D point cloud processing and spatial recognition, a series of groundbreaking methodologies have redefined the field. PointNet++[36] introduces a hierarchical neural network utilizing recursive PointNet[37] on nested input point sets, setting a new performance benchmark for 3D point cloud processing. Riding the wave of transformer success, Point Transformer leverages self-attention networks for 3D point cloud tasks like semantic scene segmentation and object categorization, outperforming prior models. Point Cloud Transformer (PCT)[38] leverages the transformer's prowess for shape classification, part and semantic segmentation, and error estimation, enhancing input embedding and achieving remarkable results.

Through the innovative fusion of convolution and Transformer, 3D Convolution-Transformer Network ([39]) attains state-of-the-art classification performance on ModelNet40. The ingenious 3D ViT, Simple3D-Former[40], bridges the gap between 2D and 3D tasks, capitalizing on the Simple3D-Former[40] architecture to tackle well-established 3D challenges. RangeViT[41], a standout innovation, outperforms existing projection-based techniques on nuScenes and SemanticKITTI, thanks to its RGB image architecture, customized convolutional stem, and pixel-wise predictions. Collectively, these cutting-edge approaches reconfigure 3D point cloud analysis, reshaping the landscape of spatial recognition and processing.

In terms of segmentation, a series of pioneering methodologies have emerged, each reshaping the field in unique ways. ViP-DeepLab[42] introduces Depth-aware Video Panoptic Segmentation, combining video panoptic segmentation and monocular depth estimation, achieving cutting-edge results and first place in multiple benchmarks. Tokens-to-Token Vision proposes an effective backbone for vision transformers out performing ResNets on ImageNet with a novel image layer wise tokens-to-tokens transformation. ConViT[43] addresses the challenge of locality in self-attention layers, exploring its role in learning and introducing a gating parameter for attention control. Visformer[44] presents a transition from Transformer to convolution-based models, unveiling the Vision-friendly Transformer architecture. SegFormer-B0[45] to SegFormer-B5[45] demonstrate superior performance and efficiency in semantic segmentation tasks. SegViT[46] leverages attention mechanisms for semantic segmentation, achieving new state-of-the-art performance on multiple datasets. TSViT[47] introduces temporal-then-spatial factorization for satellite image time series, providing an intuitive approach for SITS processing. These transformative approaches collectively redefine the boundaries of vision and image analysis, forging new horizons in research and application.

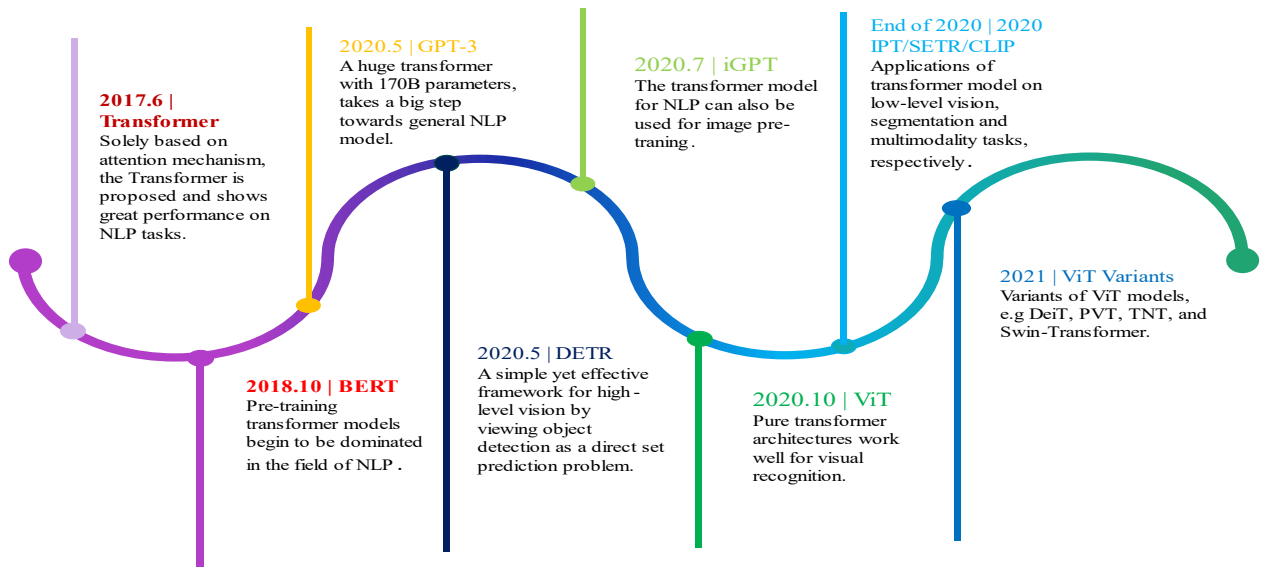

Fig. 2. Key milestones in the development of the transformer.

Fig. 4 depicts a graphical summary of ViT. Using an NLP transformer for computer vision applications necessitates a few more steps. ViTs general architecture may be divided into five major components:

- Split the image into non-overlap/overlap patches (16 × 16, 32 × 32, etc.)
- Flatten patches and generate lower-dimensional linear embeddings from the flattened patches referred to (Patch Embedding).
- Add positional embedding and class token.
- Feed the sequence of patches into the transformer layer and obtain the output (label) through a class token.
- Pass class token values to MLP head to obtain final output prediction.

**Step 1:** Given an input image of 112*112, produce 16*16 non-overlap/overlap patches. As a result, we may construct 49 patches and input them directly into the linear projection layer. Keep in mind that each patch has three color channels. The patches are sent into the projection layer to generate a long vector representation of each patch, and these representations are shown in Figure 3.

*A. Patch Embedding*

The total number of overlap/nonoverlap patches is 49, patch size with several channels is 16 × 16 × 3. The size of the long vector of each patch is 768. Overall, the patch embedding matrix is 49x196. Further, the class tokens have been added to the sequence of embedded patches and also added position Embedding. The transformer cannot maintain information without positional encoding, and accuracy is reduced to about 3%. Because of the new class token, the patch embedding size has increased to 50. Finally, the patch embeddings with positional encoding and class token are fed into the transformer layer to produce the learned class token representations. As a result, the transformer encoder layer output is 1x768 and is transferred to the MLP block to produce the final prediction.

*B. Transformer Encoder Layer*

Especially in ViTs, the most important component is the transformer encoder that contains MHSA and MLP block. The encoder layer receives combined embeddings (patch embeddings, positional embeddings, and class tokens) of shape 50 (49 patches and 1 [cls] token) ×768(16×16×3) as input. For all layers, the inputs and output of the matrix shape are 50x768 from the previous layer. In ViT In base architecture, there are 12 heads (also known as layers). Before feeding input into the MHA block, the input is normalized through the normalization layer in Figure 3. In MHA, the inputs are converted into a 50 × 2304(768 × 3) shape using a Linear layer to obtain the Query, Key, and Value matrix.

## IV. TRANSFORMER WITH CONVOLUTION

Even though vision transformers have been effectively applied to a variety of visual applications because of their capacity to capture long-range dependencies within the input, there are still performance gaps between transformers and conventional CNNs. One major cause might be a lack of capacity to retrieve local information. Aside from the previously stated ViT versions that improve locality, combining the transformer with convolution might be an easier technique to bring locality into the ordinary transformer.

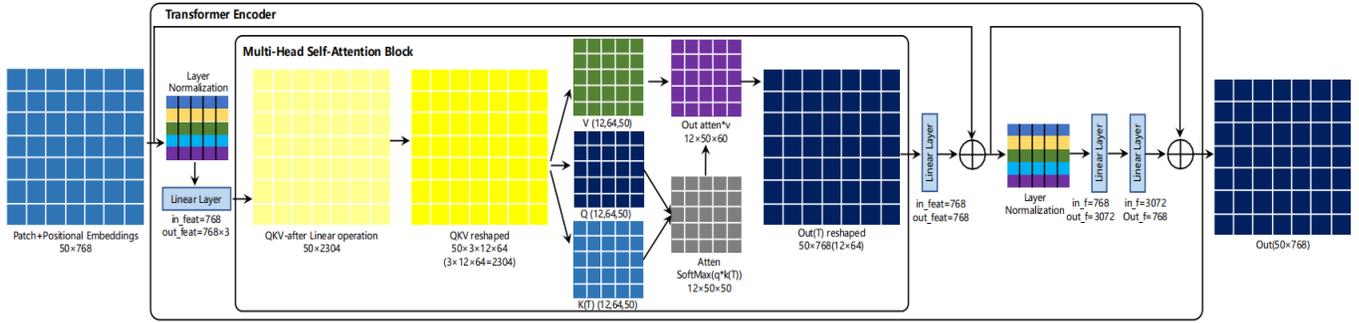

Fig. 3. Overview of transformer encoder block in vision transformer along with multi-head self-attention module

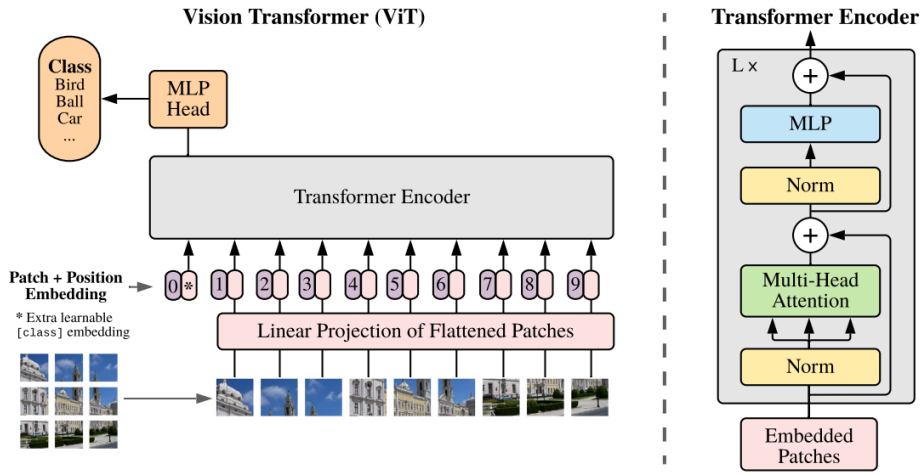

Fig. 4. Architecture of Base ViT

Several efforts attempt to enhance a traditional transformer block or self-attention layer using convolution. To exploit convolutions for fine-level feature encoding, CPVT[48] suggested a conditional positional encoding (CPE) technique that is conditioned on the immediate neighborhood of input tokens and adaptable to any input size. CvT[49], LocalViT[50], and CMT[51] investigated the possible problems of directly taking Transformer topologies from NLP and combining convolutions with transformers. In particular, the feed-forward network (FFN) in each transformer block is paired with a convolutional layer that increases token correlation. Furthermore, some studies have shown that transformer-based models are more difficult to get a favorable ability of fitting data; in other words, they are sensitive to the optimizer, hyper-parameter, and training schedule. With two distinct training settings, Visformer[44] highlighted the gap between transformers and CNNs. The first is the typical CNN setup, in which the training cycle is shorter and the data augmentation consists just of random cropping and horizontal flipping. The other is the training environment, which includes a longer training schedule and stronger data augmentation. [52] changed the early visual processing of ViT by replacing its embedding stem with a standard convolutional stem, and found that this change allows ViT to converge faster and enables the use of either AdamW or SGD without a significant drop in accuracy.

## V. APPLICATION OF VIT IN COMPUTER VISION

ViTs have been used in a variety of CV assignments with remarkable and, in some cases, cutting-edge results. Some of the major application areas are as follows:

- Image Classification
- Anomaly Detection
- Object Detection
- Image Compression
- Image Segmentation
- Video Deepfake Detection
- Cluster Analysis

According to Fig. 6, the percentage of ViTs used Image Classification, Semantic Segmentation, Object Detection, Self-Supervised Learning, Classification, Action Recognition, Instance Segmentation, Image Segmentation, Medical Image Segmentation are 8.70, 7.99, 4.58, 3.29, 2.47, 2.00, 1.76, 1.76, 1.53 respectively ViTs are commonly used in CV work. The challenges that CNNs face can be solved by ViTs. ViTs variations are used for image compression, super-resolution,

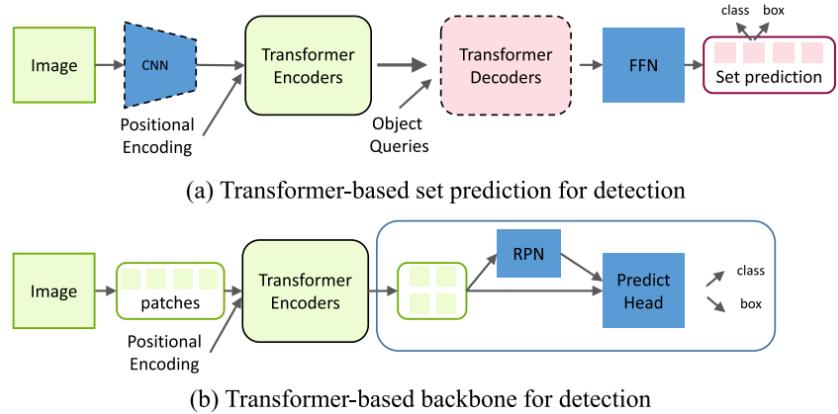

(a) Transformer-based set prediction for detection

(b) Transformer-based backbone for detection

Fig. 5. General framework of transformer-based object detection

and other applications. With the advancement in the ViTs for CV applications, a state-of-the-art survey is required which highlights the performance of ViTs for the CV tasks.

## VI. IMAGE CLASSIFICATION

The image is first separated into patches, which are then supplied linearly to the transformer encoder, where MLP, normalization, and multi-head attention are used to construct embedded patches. The MLP head predicts the output class based on embedded patches. Many academics have employed these traditional ViTs to categorize visual things. In [20], the authors proposed CrossViT-15, CrossViT-18, CrossViT-9†, CrossViT-15†, and CrossViT-18† for the image classification.

They used ImageNet1K, CIFAR10, CIFAR100, pet, crop disease, and ChestXRay8 datasets to evaluate the different variants of CrossViT[20]. They achieved 77.1% accuracy on the ImageNet1K dataset by using CrossViT-9†. Similarly, they attained 82.3% and 82.8% accuracy on the ImageNet1K dataset using CrossViT-15† and CrossViT-18† respectively. Similarly, the authors got 99.0% and 99.11% accuracy with CrossViT15 and CrossViT-18, respectively, on the CIFAR10 dataset. However, they obtained 90.77% and 91.36% accuracy on the CIFAR100 dataset using CrossViT-15 and CrossViT-18, respectively. CrossViT[20] was also employed for pet classification, crop disease classification, and chest X-ray classification by the authors. CrossViT18 provided the greatest accuracy of 95.07% for pet categorization. Similarly, they reached the best accuracy of 99.97% for crop disease classification using CrossViT-15 and CrossViT18. Furthermore, they attained the greatest accuracy of 55.94% for the chest X-ray categorization utilizing CrossViT-18. Yu et al. in [53], presented multiple instances of enhanced ViT (MIL-ViT) for fundus image classification. They used APTOS 2019 blindness detection and the 2020 retinal fundus multi-disease image dataset (RFMiD2020). MIL-VT[53] gave an accuracy of 97.9% on the APTOS2019 dataset and 95.9% on the RFMiD2020 dataset. Xue et al. [54] proposed deep hierarchical ViT (DHViT) for the hyperspectral and light detection and ranging (LiDAR) data classification. The authors used Trento, Houston 2013, and Houston 2018 datasets and obtained an accuracy of 99.58%, 99.55%, and 96.40%, respectively.

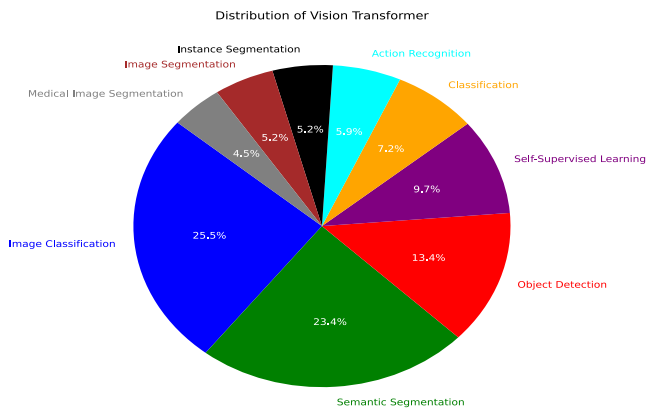

Fig. 6. Use of ViTs for CV Applications

## VII. 3D OBJECT CLASSIFICATION

Object classification in 3d images is similar to 2d images. They can be further narrowed down to two parts. One for global level classification another is for local level classification. Global level: At the global scale, several methods have integrated the attention mechanism into various parts of the network, with different inputs and position embeddings. Attentional Shape ContextNet [55] was one of the early adopters of self-attention for point cloud recognition. To learn shape context, the self-attention module is used to select contextual regions, aggregate and transform features. This is carried out by replacing hand- designed bin partitioning and pooling with a weighted sum aggregation function with input learned by self-attention applied on all the data. Adaptive Wavelet Transformer [56] performs multiresolution analysis within the neural network to generate visual representation decomposition using the lifting scheme technique. The generated approximation and detail components capture geometric information that is of interest for downstream tasks.

DTNet [60] aggregates point-wise and channel-wise multi-head self-attention models to learn contextual dependencies from the

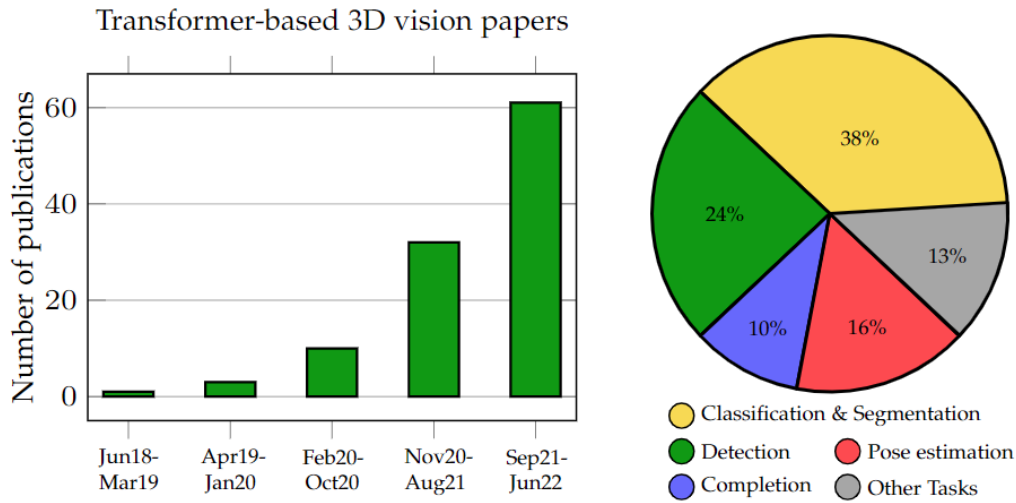

Fig. 7. A nice illustration provided by the authors [95] to showcase the applications of transformer

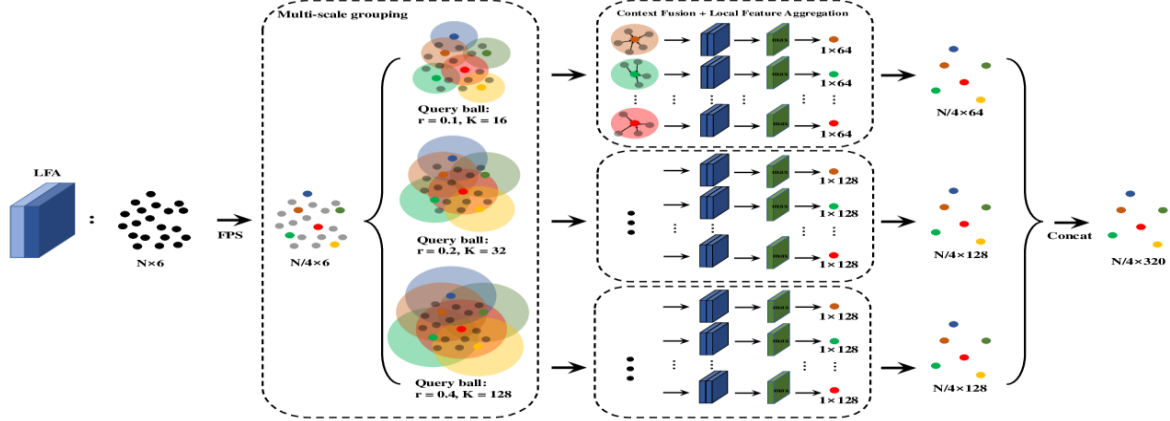

Fig. 8. Multi-Scale Local Feature Aggregating (LFA)[46] has three key steps Multi-scale Grouping (Left), Context Fusion (right) and Local Feature Aggregation(right).

A transformer is then used to pay different attention to features from approximation and detail components and to fuse them with the original input shape features. TransPCNet [57] aggregates features using a feature embedding module, feeds them into separable convolution layers with a kernel size of 1, and then uses an attention module to learn features to detect defects in sewers represented by 3D point clouds. Other methods propose variations of the attention module. Yang et al. [58] develop Point Attention Transformers (PATs) by applying an attention module to a point cloud represented by both absolute and relative position embeddings. The attention module uses a multi-head attention design with group attention, which is similar to depth wise separable convolution [59], in addition to channel shuffle. Point Cloud Transformer (PCT) [58] applies offset-attention to the input point embedding. The offset-attention layer calculates the element-wise difference between the self-attention features and the input features. It also uses a neighbor embedding by sampling and grouping neighboring points for better local feature representation.

position and channel. Some methods focus on pre-training the transformer by masking parts of the input. Point-BERT [61] first partitions the input point cloud into point patches, inspired by Vision Transformers [62], and uses a mini-Pointnet [63] to generate a sequence of point embeddings. The point embeddings are then used as an input to a transformer encoder, which is pre-trained by masking some point embeddings with a mask token, similar to [13].

The tokens are obtained using a pre-learned point Tokenizer that converts the point embeddings into discrete point tokens. Similarly, Pang et al. [64] divide the input point cloud into patches and randomly mask them during pre-training. A transformer-based autoencoder is used to retrieve the masked point patches by learning high-level latent information from unmasked point patches. At the local scale, Point Transformer [65] applies self-attention in the local neighborhood of each data point. A point transformer block consists of the attention layer, linear projections, and residual connection. Additionally, instead of using the 3D point coordinates as position encoding, an encoding function is used with linear layers and ReLU nonlinearity. To increase the receptive field of the proposed transformer architecture, transition down layers is introduced, as well as transition up to retrieve the original data size. 3DCTN [39] proposes to combine graph convolution layers with transformers. The former learns local features efficiently,

Detection. As a pioneer in transformer-based detection methods, the detection transformer (DETR) proposed by Carion et al. [66]redesigns the framework of object detection. DETR, a simple and fully end-to-end object detector, approaches the challenge of object identification as an intuitive set prediction problem, omitting standard hand-crafted whereas the latter is capable of learning global context. Taking the point cloud with normal as input, In this way the authors tries to combine both approaches together in a same network.

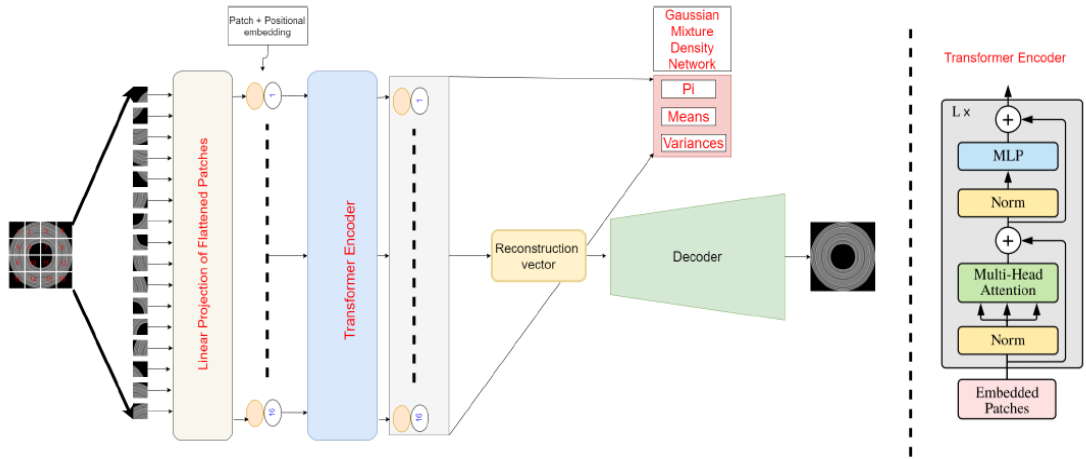

Fig. 9. Network architecture of VT-ADL [86] Used for Anomaly Detection and localization.

## VIII. GENERIC OBJECT DETECTION

Traditional object detectors are mostly based on CNNs, however, transformer-based object detection has lately gained popularity because of its favorable capabilities. Some object detection approaches have sought to use the transformer's self-attention process before improving certain modules for newer detectors, such as the feature fusion module and prediction head. Transformer-based object identification methods are broadly classified into two groups: transformer-based set prediction methods and transformer-based backbone methods. When compared to CNN-based detectors, transformer-based approaches outperformed them in terms of both accuracy and operating speed. Transformer-Based Set Prediction for components such as anchor creation and non-maximum suppression (NMS) post-processing. DETR begins with a CNN backbone to extract features from the input picture, as illustrated in Figure 6. Fixed positional encodings are added to the flattened features before they are supplied into the encoder-decoder transformer to supplement the image features with location information.

## IX. SEGMENTATION

[67]replaced the conv-encoder with a pure transformer and introduced a sequence-to-sequence technique Three distinct decoders are intended to execute pixel-wise segmentation utilizing progressive sampling, multi-level feature aggregation, and a naïve up-sampling technique. [68] presented a convolution-free end-to-end trainable encoder and decoder that captures contextual information. The encoding portion is built on standard ViT and depends on the encoding of patches.

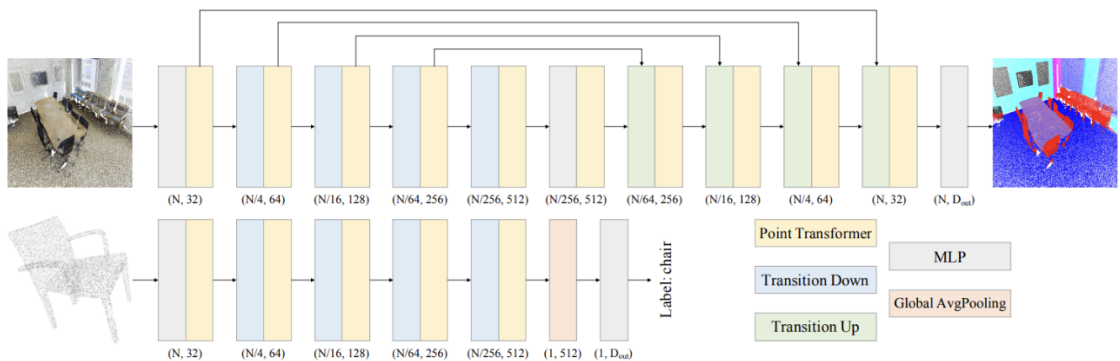

Fig. 10. Point Transformer Architecture[65]. Where Blocks architecture are given by colors

Further, a point-wise linear decoder is applied and decoded with a mask transformer. [45] presented SegFormer, a simple yet powerful method with lightweight MLP decoders. An encoder is based on a hierarchical structure that gives multi-scale features and does not require any positional encoding scheme. SegFormer gets rid of complex decoders combining local attention and global attention to gender representation. [67] proposed a complete transformer network that uses a pyramid group transformer[69] encoder to advance learned hierarchical features while lowering the computational cost of regular ViT. In addition, the feature pyramid transformer combined information from the geographical and semantic levels of the pyramid group transformer.

## X. 3D Image Processing

While Transformers has seen great success in 2D image domain. One of the particular challenges is to encode positional information to the attention mechanism. Because 2D images are a projection of 3D space on a 2d Image. So generally, when we project some higher dimensional information to a lower dimensional space, we lose some information. So, it's an intuition that the researcher believes that using transformer in the original space should contain much more information. That would give the transformer to learn more semantic information from the data. Unlike 2d images, 3d data contain positional information in 3d coordinate system. So, in order these data to be used with the transformer some preprocessing is needed Fig. 10 shows the application of ViT in 3d image processing. There are several preprocessing techniques that are widely used in the literature like random sampling, farthest point sampling and k Nearest Neighbours (kNN) sampling to name a few. Generally, these samplings are done before training process but some authors have merged the preprocessing during training lead to overhead during training. Since Transformers can work with arbitrary length of input data. One proposed solution is to process the point clouds from 3d representation directly with transformer to convert the input data to a regular grid. Now the representation of spaced voxels allows similar representations of the object irrespective of its size. In the following section we will highlight some leading research of using Vision Transformer for object classification in 3d image.

Local level Point clouds are widely used to represent 3D objects in various computer vision applications. The success of deep learning on 2D images has inspired researchers to extend deep learning techniques to point clouds. One of the challenges in applying deep learning to point clouds is that they lack a structured representation, unlike images and videos. Therefore, researchers have proposed various methods to process point clouds with deep learning techniques. A common approach to process point clouds is to use convolutional neural networks (CNNs) adapted for point clouds. These networks learn local features by aggregating information from neighboring points. One of the early approaches to use CNNs for point clouds is PointNet [70] PointNet[71] uses max pooling to aggregate features across all points and then applies a multi-layerperceptron (MLP) to learn global features. The main limitation of PointNet[70] is that it cannot capture local structures of point clouds, such as edges and corners. To overcome this limitation, PointNet++ [70], [71] extends PointNet with a hierarchical grouping strategy that allows the network to learn local structures at different scales. Another approach to process point clouds is to use graph neural networks (GNNs). GNNs generalize the convolution operation to arbitrary graphs, allowing them to handle point clouds with irregular structures. PointRGCN [72] proposes a graph convolutional network that uses a multi-scale neighborhood aggregation strategy to learn local features. DGCNN [73] uses a dynamic graph construction algorithm to build a k-nearest neighbor graph and applies graph convolutional layers to learn local features. The main limitation of GNNs is that they require the graph structure to be defined beforehand, which can be challenging for complex point clouds. Recently, researchers have proposed to use attention mechanisms for processing point clouds. Attention mechanisms allow networks to selectively attend to relevant features, which can be beneficial for point clouds where the relevant features can be highly variable across different regions. Point Transformer [74] applies self-attention in the local neighborhood of each data point. A point transformer block consists of the attention layer, linear projections, and residual connection. Additionally, instead of using the 3D point coordinates as position encoding, an encoding function is used with linear layers and ReLU nonlinearity. To increase the receptive field of the proposed transformer architecture, transition down layers is introduced, as well as transition up to retrieve the original data size. Another approach that combines graph convolution layers with transformers is 3DCTN [75]. The former learns local features efficiently, whereas the latter is capable of learning global context. Taking the point cloud with normal as input, the network consists of two modules that down sampled the point set, with each module having two blocks: the first block is a local feature aggregation module using a graph convolution, and the second block is a global feature learning module using a transformer consisting of offset-attention and vector attention. LFT-Net [76] proposes a local feature transformer network that uses self-attention to learn features of point clouds. It also introduces a Trans-pooling layer that aggregates local features to reduce the feature size. The network consists of two stages, the first stage applies a series of multi-scale neighborhood aggregations to learn local global features. features, and the second stage applies self-attention to learn

## XI. ViTs in Image Denoising

Image denoising is a crucial task in digital image processing, with various applications in fields such as medical imaging, astronomy, and computer vision. Traditional denoising techniques relied on hand-crafted features and mathematical models, which required expert knowledge and assumptions about the image and noise characteristics. However, with recent advances in machine learning, there has been a shift towards data-driven approaches for denoising. Convolutional neural networks (CNNs) have shown great promise in image denoising due to their ability to learn complex representations directly from data. One popular CNN-based approach for denoising is the Variational Autoencoder (VAE) with an added component known as the Variational Information Theoretic (Vit) loss function. The Vit loss function utilizes an information-theoretic measure called Mutual Information (MI) to better model the underlying image distribution and noise characteristics, resulting in improved denoising performance. Studies have shown that the use of Vit in image denoising outperforms traditional hand-crafted approaches, with better results in terms of signal-to-noise ratio (SNR) and peak signal-to-noise ratio (PSNR). For instance, in a recent study by Wang et al. (2021)[77], a Vit-based denoising method was shown to achieve state-of-the-art performance on the widely-used Berkeley Segmentation Dataset (BSD). Another study by Liu et al. (2020) showed that a Vit-based approach outperformed traditional denoising methods on both synthetic and real-world noise. In this section, we explore recent approaches for image denoising, with a particular focus on the TED-Net[77]. Traditional Vision Transformers (ViTs) consist of an encoder and decoder layer, with a transformer head used for feature extraction from the image. In contrast, the authors of TED-Net split the network into multiple blocks, each comprising two distinct parts. The first block, known as the Transformer-block (TB), uses a traditional transformer architecture (encoder, decoder block) placed between two Tokens to Tokens blocks. Additionally, multi-layer perceptron's and residual[78] connections are incorporated for feature aggregation. The output size of the TB block is the same as the input token. In subsequent layers, the authors introduce the **Token-To-Token** Dilated Block **(T2TD),** which takes the output from the previous stage. The T2TD block overcomes the limitation of simple tokenization of the image in ViT by adopting a cascade tokenization procedure. Tokens are transposed to T' and reshaped into $I \in R^{b \times n \times d}$ (I= input tokens, $c=d, h = w = n$). A soft-split with dilation is performed, reducing the four-dimensional feature maps into three dimensions by combining several neighboring tokens into one. After compressing the feature map, a cyclic shift is performed to integrate more information into the model. Finally, an inverse cyclic shift is performed in the symmetric decoder network to avoid pixel shift in the final denoising layers. Then they have combined the noise information from the model with the original image to get the final output. The proposed TED-Net[79] has demonstrated improved performance in image denoising compared to previous approaches. For example, in a recent study by the authors, the proposed TED-Net[79] outperformed other state-of-the-art methods in terms of peak signal-to-noise ratio (PSNR) and structural similarity index (SSIM). Overall, the results suggest that the use of a split network with the TB and T2TD blocks can effectively capture and integrate features for image denoising, yielding improved performance over traditional ViT approaches. The proposed approach was evaluated using the publicly available 2016 NIH-AAPM-Mayo clinic LDCT grand challenge dataset. The authors used the patient L506 data for evaluation and the remaining nine patients for model training. The authors applied various augmentation techniques during training to improve the robustness of the model. The proposed approach was compared against several state-of-the-art baseline algorithms, including RED-CNN [80], WGAN-VGG [81], [82], MAP-NN [82], and AD-NET [83]. The experimental results showed that the proposed approach outperformed the low-dose CT denoising models while retaining the original details of the target image. Furthermore, the proposed approach achieved similar performance to high-performance gray image denoising models.

## XII. ViTs in Anomaly Detection

Anomaly detection is a critical task across various domains, such as healthcare, finance, and security, as it involves identifying unusual patterns or events within data that could indicate malfunctions or abnormalities. In recent times, there has been a growing interest in exploring the capabilities of Vision Transformers (ViTs)[62] for anomaly detection. ViTs have demonstrated remarkable potential due to their attention mechanism's ability to capture intricate details within images and enable efficient processing of large datasets. In this section, we delve into recent studies that have investigated the utilization of ViTs in anomaly detection, highlighting both the strengths and limitations of this approach. ViT's attention mechanism stands out as a key feature, enabling the model to focus on relevant parts of the image and capture nuanced information that might be missed by traditional methods. Moreover, the scalability of ViTs makes them well-suited for processing extensive datasets, contributing to their applicability in real-world scenarios. However, there are challenges associated with adapting ViTs to unsupervised anomaly detection and addressing issues related to imbalanced data distributions. One promising approach in this field is the AnoVit[84] architecture, which employs an image reconstruction-based framework. AnoVit's[84] architecture consists of an encoder and decoder pipeline. The encoder employs ViT to extract latent features from input images, while the decoder uses these features to reconstruct the image in a higher-dimensional space. Notably, the proposed encoder in AnoVit[84] employs a unique rearrangement of the feature map in three dimensions, allowing for direct feature map utilization without additional layers in the decoder block. This preserves spatial information between patches and contributes to improved reconstruction accuracy.

To train the AnoVit[84] model for anomaly detection, the network is trained on normal images and their corresponding reconstructions. A key principle is that the network learns rich features suitable for reconstructing normal images during training. When presented with anomalous data, the network struggles to reconstruct the image accurately, resulting in a higher reconstruction error. The anomaly score is computed based on this error, and a threshold is employed for classification. Evaluation of AnoVit[84] on diverse datasets, including challenging benchmarks like MVTec-AD[85], showcases its effectiveness. Comparisons with traditional methods like convolutional neural networks (CNNs) and hybrid CNN-ViT models demonstrate AnoVit's[84] superiority in detecting anomalies, as evidenced by metrics such as AUC, precision, and recall. Furthermore, AnoVit[84] exhibits robust generalization across various domains, from medical images to satellite imagery, indicating its potential for broad applications.

Another noteworthy approach, VT-ADL[86], employs ViTs for anomaly detection with a distinct perspective. This framework utilizes a specialized encoder-decoder architecture and introduces multiple training objectives. The image is divided into patches and encoded using a conventional ViT, followed by decoder training to learn features representative of normal images. The authors enhance this approach by incorporating a Gaussian Mixture density network to model the distribution of transformer-extracted features. This architecture enables automatic localization of anomalies, facilitated by positional encoding. The encoder in VT-ADL[86] combines multi-layer perceptron's (MLPs) with transformers. Layer normalization, residual connections, and strategic dropout elimination contribute to model stability. The decoder process involves projecting encoded patches to reconstruction vectors through learnable projection matrices. Reconstruction into images is achieved using transposed convolution layers with appropriate activation functions. VT-ADL[86] employs a dual-objective training strategy. The first objective focuses on image reconstruction, employing Mean Squared Error (MSE) and Structural Similarity Index (SSIM) loss functions. The second objective pertains to the Gaussian Mixture density network, aiming to minimize negative conditional log-likelihood. This multi-objective training strategy empowers VT-ADL[86] to achieve state-of-the-art performance in terms of anomaly detection and localization. The evaluation on benchmark datasets like MVTec AD[85], BTAD, and MNIST[87] underscores VT-ADL's[86] efficacy and superiority over existing methods

XIII. OTHER TASKS

In recent years, vision transformers used in so many computer vision fields rather than not just for object detection or semantic segmentation, or classifications. It's also used in Image compression, 3D points clouds, Image super-resolution, and image denoising either. In the next phase of this paper, We will briefly discuss them to get some basic ideas.

*A. Image Compression*

In recent years, the field of learning-based image reduction has garnered substantial attention as a promising avenue for efficient image compression. Various architectures based on Convolutional Neural Networks (CNNs) have demonstrated their efficacy in achieving lossy picture compression through learning processes. As the capabilities of transformers, particularly Vision Transformers (ViTs), have advanced, these models have also been explored for learning-based image reduction tasks. In a notable study by the authors of [88], ViTs were employed to enhance the entropy module of the Balle 2018 model, leading to the emergence of a novel architecture termed Entroformer. This unique hybrid model combined the strengths of transformers and entropy modeling to optimize compression performance. The utilization of transformers within the entropy module, coined as Entroformer, facilitated the efficient capture of long-range dependencies within the probability distribution estimation process. One pivotal aspect of Entroformer's contribution was its adeptness in effectively capturing intricate relationships across distant regions of the image. By leveraging the inherent capabilities of transformers, the model exhibited enhanced performance in encoding probability distributions, which is crucial for successful image compression. This achievement is particularly significant given that long-range relationships are challenging to capture using conventional methods. To validate the effectiveness of Entroformer, the researchers conducted experiments on the Kodak dataset, a well-known benchmark for image compression evaluation. The performance metrics, namely the average Peak Signal-to-Noise Ratio (PSNR) and the Multi-Scale Structural Similarity (MS-SSIM), were employed to assess the quality of the compressed images produced by Entroformer. When the model was fine-tuned for the Mean Squared Error (MSE) loss function, the achieved PSNR and MS-SSIM values were 27.63 dB and 0.90132, respectively.

*B. 3D Point Clouds*

The goal of 3D point cloud data collection is to gather data in a place with rich geometric information, shape knowledge, and scale knowledge. Each data point in a point cloud is represented as a 3D object or form with Cartesian coordinates (X, Y, Z). Data from 3D sensors and devices, as well as photogrammetry software, is used to capture point data. Due to long-term dependencies, self-attention-based point cloud approaches have recently gained substantial relevance, particularly this year. Different point cloud methodologies have been developed, and SOTA outcomes have been obtained when compared to deep learning methods. Recent advancements in 3-D sensing technology have driven a shift towards 3-D computer vision in various fields, such as augmented reality, 3-D reconstruction, and autonomous systems. While 3-D point cloud data is readily available, its unstructured nature poses challenges for processing. Traditional convolutional neural networks (CNNs) are not directly applicable. To address this, [89] introduce the Space-Cover Convolutional Neural Network (SC-CNN), which leverages Space-Cover Convolution (SC-Conv) to dynamically learn spatial geometry in local point cloud subsets. SC-CNN effectively captures complex shape information and achieves superior results in point cloud classification, part segmentation, and scene segmentation. This survey paper contributes a novel approach to efficient 3-D shape perception. Additionally The rapid development of 3-D sensing technology has transformed various fields, including augmented reality, 3-D reconstruction, and autonomous systems. However, processing 3-D point cloud data remains a challenge. To address this, [90]introduce SC-CNN, a Space-Cover Convolutional Neural Network that efficiently learns spatial geometry in local point cloud subsets. SC-CNN excels in point cloud classification, part segmentation, and scene segmentation, offering a promising solution for advanced 3-D shape perception in a concise manner. However these papers explored network configurations without any presence of Transformer. There are recent research exploring the use of Transformers in processing 3D point clouds. [32] built an amazing layer to act as a backbone of intensive prediction stacks and scene interpretation. A suggested layer is permutation and cardinality insensitive, making it intrinsically ideal for processing point cloud data. Using the suggested layer, they built a point transformer that can reliably handle point cloud data. Long-distance connections are difficult to capture in CNN networks; the transformer overcomes this problem with a self-attention mechanism.

XIV. OPEN RESEARCH CHALLENGES AND FUTURE WORKS

Despite demonstrating good results for several image coding and CV challenges. In addition to high computational costs, extensive training datasets, neural architecture search, transformer interpretability, and economical hardware designs, ViT's implementation confronts other hurdles. The goal of this section is to describe ViT's difficulties and future directions.

*A. High computational cost*

ViT-based models include millions of parameters. To train these models, computers with great computing power are required. These high-performance processors raise the computational cost of ViTs due to their high cost. ViT outperforms CNN; nonetheless, its computing cost is significantly larger. One of the most difficult tasks for academics is lowering the computing cost of ViTs.

*B. Large training dataset*

The training of ViTs necessitates a vast amount of data. ViTs perform badly with a limited training dataset. ViT trained on the ImageNet1K dataset underperforms ResNet, whereas ViT trained on the ImageNet21K dataset outperforms ResNet.

*C. Neural architecture search (NAS)*

There has been a lot of research on NAS for CNNs. NAS, on the other hand, has not yet been investigated for ViTs. The NAS ViTs exploration provides a fresh avenue for young investigators.

*D. Efficient hardware design*

Large-scale ViTs networks may be inappropriate for edge devices and resource-constrained environments such as the internet of things due to power and computation needs (IoT).

XV. CONLUSION

Because of the benefit of the self-attention mechanism and the creation of a long-term connection, ViTs are gaining popularity and producing increasingly impressive outcomes in the domain of CV. In this paper, we first introduced the core ideas of vision transformers before reviewing modern ViT approaches. We also discuss about ViT is being used in all domain of image processing, Second, we highlight important work and create a chronology of ViTs in many fields (for example, classification, segmentation, point cloud, and object detection). Finally, we compare several ViTs and CNN approaches in terms of accuracy on an ImageNet dataset. After the advent of ViT we see the SOTA metrics is being pushed forward in almost all types of public dataset. This goes to show the high efficiency of ViT. Although Transformers are originated from language domain but is highly successful in Computer Vision tasks as well highlighting generalization capabilities of this architecture. Overall, ViT performs quite well and combining the self-attention mechanism with CNN yields exceptional results. However, all these benefits come at a high cost of training time. Vision Transformers are notoriously hard to train and requires significantly larger training time compared to CNN. And Transformers requires more data samples making CNN still the better choice in certain scenario. In this paper we try to focus on different approach researchers are using for different problem set and how they are being adopted in Computer Vision problems. Finally, vision transformers are still an active area of research, where the researchers are constantly trying to address all the shortcomings of the transformer.

| | |
|---|---|
| | *Pattern Recognition*, 2021. doi: 10.1109/CVPR46437.2021.00542. |
| [93] | K. Lin, L. Wang, and Z. Liu, "End-to-End Human Pose and Mesh Reconstruction with Transformers," in *Proceedings of the IEEE Computer Society Conference on Computer Vision and Pattern Recognition*, 2021. doi: 10.1109/CVPR46437.2021.00199. |